\documentclass[lettersize,journal]{IEEEtran}
\usepackage{float} 
\usepackage{amsmath,amsfonts}
\usepackage{algorithmic}
\usepackage{algorithm}
\usepackage{array}
\usepackage[caption=false,font=normalsize]{subfig}
\usepackage{textcomp}
\usepackage{stfloats}
\usepackage{url}
\usepackage{verbatim}
\usepackage{graphicx}
\usepackage{cite}
\usepackage{multirow}
\usepackage{tabularx,booktabs}
\newcolumntype{C}{>{\centering\arraybackslash}X} 
\usepackage[dvipsnames]{xcolor}
\newcommand{\best}[1]{\textcolor[rgb]{1,0,0}{\textbf{#1}}}
\newcommand{\secondbest}[1]{\textcolor[rgb]{0,0,1}{\underline{#1}}}
\usepackage{longtable}
\usepackage{makecell, cellspace, caption}
\usepackage{array}
\usepackage{subcaption}
\usepackage{cuted}
\usepackage{multirow}
\usepackage[utf8]{inputenc}
\usepackage{tabularx}
\usepackage{array}
\usepackage{amssymb}
\usepackage{makecell}
\usepackage[margin=2cm]{geometry}
\usepackage{adjustbox}
\usepackage{multirow}
\usepackage{tabulary,booktabs}
\usepackage{ragged2e}
\usepackage{amsmath}
\usepackage{booktabs}
\usepackage{array}
\newcolumntype{L}{>{$}l<{$}}
\newcolumntype{C}{>{$}c<{$}}
\newcolumntype{R}{>{$}r<{$}}
\newcolumntype{P}{>{}p{5.5em}<{}}
\newcolumntype{F}{>{$}p{3.1em}<{$}}
\newcolumntype{A}{>{$}p{2.5em}<{$}}
\newcolumntype{B}{>{$}p{3em}<{$}}
\newcolumntype{D}{>{$}p{2em}<{$}}
\newcolumntype{E}{>{$}p{1.5em}<{$}}
\newcolumntype{G}{>{$}p{1em}<{$}}

\usepackage{amssymb}
\usepackage{pifont}
\usepackage{graphicx}
\usepackage{hyperref}
\hypersetup{colorlinks,allcolors=blue}
\usepackage{flushend}
\usepackage{diagbox}
\usepackage{mathtools}
\begin{document}

\title{VSANet: View-aware Sparse Attention Network for Light Field Image Denoising}

\author{Gargi Panda, Soumitra Kundu, Saumik Bhattacharya, Aurobinda Routray
\thanks{Gargi Panda and Aurobinda Routray are with the Department of EE, IIT Kharagpur, India
(email: pandagargi@gmail.com; aroutray@ee.iitkgp.ac.in)}
\thanks{Soumitra Kundu is with the Rekhi Centre of Excellence for the Science of Happiness, IIT Kharagpur, India (e-mail: soumitra2012.kbc@gmail.com).}
\thanks{Saumik Bhattacharya is with the Department of E\&ECE, IIT Kharagpur, India
(email: saumik@ece.iitkgp.ac.in)}
}

\markboth{Journal Submission}%
{Shell \MakeLowercase{\textit{et al.}}: A Sample Article Using IEEEtran.cls for IEEE Journals}


\maketitle
\begin{abstract}
Light field (LF) image denoising is challenging due to the high-dimensional structure of LF data. While noise is independent across sub-aperture images, scene content exhibits strong cross-view correlations. We introduce VSANet, a view-aware sparse attention network for LF denoising. Specifically, we propose a view-aware sparse attention (VSA) block that represents the 4D LF feature map as a unified spatial-angular token space and performs cross-view aggregation via locality-sensitive hashing-based sparse attention. This enables global feature interactions with linear complexity, effectively exploiting LF correlations across views and spatial locations. In addition, we design a feature refinement (FR) block to emphasize informative features in spatial, angular, and epipolar subspaces. The VSA and FR blocks are integrated within a sequential attention refinement module, forming the core of VSANet. Experiments demonstrate VSANet outperforms state-of-the-art LF denoising methods.
\end{abstract}
\begin{IEEEkeywords}
Light field image denoising, sparse attention, non-local similarity, epipolar geometry.
\end{IEEEkeywords}

\section{Introduction}
\label{sec:intro}

Light field (LF) imaging captures both spatial and angular information of a scene by simultaneously recording light rays from multiple viewpoints~\cite{plenoptic,lf}.
The resulting four-dimensional (4D) representation, comprising $S\!\times\!T$
sub-aperture images (SAIs) of spatial resolution $H\!\times\!W$, enables numerous applications including depth estimation \cite{depth1}, refocusing \cite{refocusing1}, and view synthesis \cite{view1}.
However, LF acquisition systems are often affected by photon-limited noise, particularly under low-light or high-speed imaging conditions, which degrades image quality and subsequent LF processing tasks \cite{apa}. Consequently, LF denoising has become an essential preprocessing step in LF imaging pipelines \cite{imp1,imp3}.

Unlike single-image denoising, LF denoising should exploit the fact that noise is generally independent across views, whereas the underlying scene content exhibits strong spatial-angular correlations \cite{lfbm5d,apa,anisotropic,handle_lf}. Traditional methods, such as LFBM5D \cite{lfbm5d}, leverage this property through collaborative filtering of similar patches collected from neighboring views. However, their dependence on hand-crafted priors limits their adaptability to complex LF structures. Recent deep learning (DL) approaches improve denoising performance by incorporating spatial-angular regularization or epipolar geometry-guided processing \cite{drlf,pfe,msp,hlfrn}. Despite their effectiveness, these methods rely on predefined geometric decompositions and are therefore limited in modeling long-range dependencies across the complete 4D LF. Moreover, transformer-based LF methods developed for other LF tasks commonly partition LF data into isolated 2D subspaces, preventing direct interaction among all spatial-angular positions  \cite{lfsr1,EPIT,lgfn,many2many}. Additional discussion of related LF methods is provided in the supplementary material.

\begin{figure*} 
    \centering
  \includegraphics[width=\linewidth]{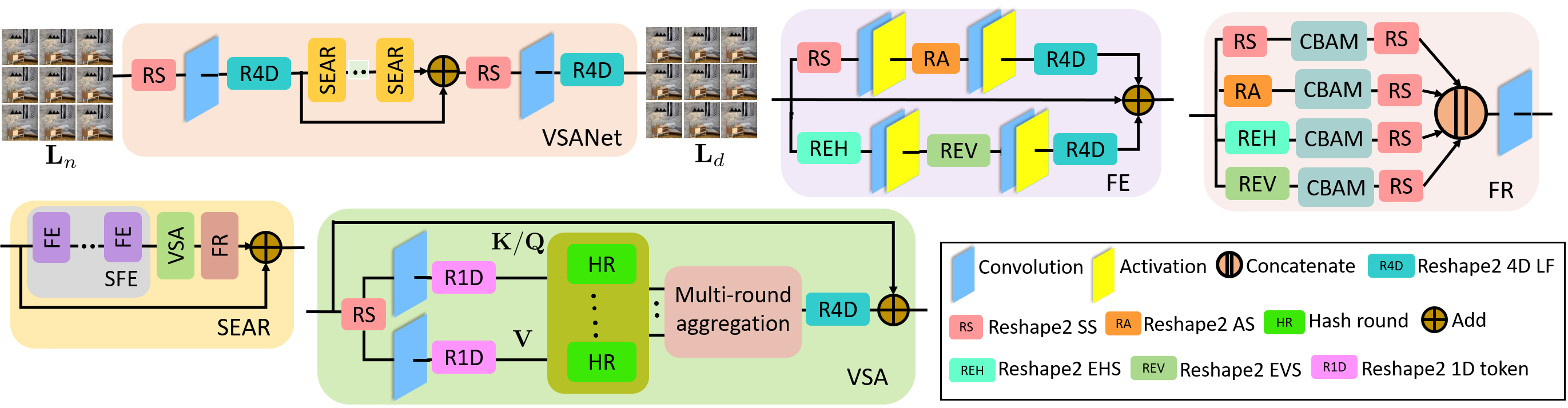}
  
  \caption{Architecture of proposed VSANet. Given the noisy LF image $\mathbf{L}_n$, VSANet generates the denoised image $\mathbf{L}_d$.}
  
  \label{fig:arch} 
\end{figure*}

To address these limitations, we propose VSANet, a view-aware sparse attention network for LF image denoising. The core of VSANet is a novel view-aware sparse attention (VSA) block that represents the entire 4D LF feature map as a unified token sequence, enabling global spatial-angular interaction within a single attention space. To maintain computational efficiency, locality-sensitive hashing (LSH) \cite{lsh,reformer} is employed to group similar tokens into sparse attention buckets, facilitating implicit non-local aggregation across views with linear complexity. In addition, a feature refinement (FR) block is introduced to enhance discriminative representations in the spatial, angular, and epipolar subspaces. The VSA and FR blocks are integrated into a sequential extraction attention refinement (SEAR) module, which serves as the fundamental building block of VSANet. To the best of our knowledge, VSANet is the first LF framework that performs attention over the entire 4D LF representation. Extensive experiments on two benchmark LF datasets under three noise levels demonstrate that VSANet outperforms existing state-of-the-art (SOTA) methods.

\section{Proposed Method}
\label{sec:method}

\subsection{Problem Formulation}
\label{subsec:problem}

A 4D LF image comprises $S\!\times\!T$ SAIs,
each with spatial resolution $\!H\!\times\!W$.
We denote the noisy LF as
$\mathbf{L}_n \in \mathbb{R}^{S\times T\times H\times W\times 3}$,
where the additive noise is assumed to be independent and identically distributed (i.i.d.) across views and spatially uncorrelated. Although the noise is view-independent, the underlying scene content exhibits strong spatial-angular correlations governed by the LF geometry~\cite{apa}. The objective of LF denoising is to estimate the denoised LF
$\mathbf{L}_d \in \mathbb{R}^{S\times T\times H\times W\times 3}$ from the noisy observation $L_n$. To this end, we propose VSANet that exploits both inter-view and intra-view correlations by performing efficient non-local aggregation over the entire 4D LF representation.


\subsection{Network Architecture}
\label{subsec:arch}

We propose VSANet, whose architecture is illustrated in Fig.~\ref{fig:arch}. The network exploits four complementary LF subspaces to model distinct spatial-angular dependencies: spatial subspace (SS, $\mathbb{R}^{S\cdot T\times H\times W\times 3}$), angular subspace (AS, $\mathbb{R}^{H\cdot W\times S\times T\times 3}$), horizontal epipolar subspace (EHS, $\mathbb{R}^{S\cdot H\times T\times W\times 3}$), vertical epipolar subspace (EVS, $\mathbb{R}^{T\cdot W\times S\times H\times 3}$).
Given a noisy LF $\mathbf{L}_n$, we first perform shallow feature extraction in the SS using a convolution layer with feature dimension $C$, capturing local intra-view spatial information. The resulting feature is then reshaped back to the 4D LF representation using R4D operation and processed through a cascade of $G$ stacked SEAR modules with a global residual connection. Finally, a convolution layer is applied in the SS, and the output is reshaped to 4D structure to obtain the denoised LF $\mathbf{L}_d$.

The core of the network relies on the three sequential stages embedded within each SEAR module: sequential feature extraction (SFE), view-aware sparse attention, and feature refinement. In the SFE stage, $F$ residual feature extraction (FE) blocks are cascaded to capture local LF structures. Each FE block contains two parallel branches. The spatial-angular branch operates in the SS and AS domains using convolution and activation layers to exploit spatial and angular correlations. The epipolar branch operates in the EHS and EVS domains to capture geometric consistency encoded in epipolar structures. The features extracted from the two branches are fused and subsequently passed to the VSA block for global cross-view non-local aggregation. The detailed formulation of the VSA block is presented in Section~\ref{subsec:vsa}. Following VSA, the FR block employs convolutional block attention modules (CBAMs)~\cite{cbam} in the SS, AS, EHS, and EVS domains. CBAMs apply channel and spatial attentions sequentially to enhance discriminative feature representations. The resulting features are concatenated and fused through a convolution layer before being forwarded to the next SEAR module.

\begin{table*}[t!]
	\centering
	\caption{Performance comparison with the SOTA methods. Mean $\pm$ SD of PSNR and SSIM are reported. We highlight the best performance in \best{red} color, and underline the second-best performance in \secondbest{blue} color.}
	\begin{tabular}{rccccccr}
		\toprule
		\multirow{3}{*}{\makecell{Method\\ (Publication)}}
		& \multicolumn{2}{c}{$\sigma=10$} 
		& \multicolumn{2}{c}{$\sigma=20$} 
		& \multicolumn{2}{c}{$\sigma=50$} 
		& \multirow{3}{*}{\makecell{\# Params(K)\;/\; \\ Runtime(s)}} \\
		
		\cmidrule(lr){2-3} \cmidrule(lr){4-5} \cmidrule(lr){6-7}
		& \makecell{STFLytro\\ PSNR \;/\;SSIM} & \makecell{EPFL\\ PSNR \;/\;SSIM} & \makecell{STFLytro\\ PSNR \;/\;SSIM} & \makecell{EPFL\\ PSNR \;/\;SSIM} & \makecell{STFLytro\\ PSNR \;/\;SSIM} & \makecell{EPFL\\ PSNR \;/\;SSIM} & \\
		
		\toprule
		\makecell{LFBM5D \\(MMSP'17) }
		& \makecell{39.61$\pm$\best{1.37}\;/\\0.9436$\pm$0.0444}
		& \makecell{37.74$\pm$\best{0.92}\;/\\0.9661$\pm$0.0126}
		& \makecell{34.23$\pm$\best{1.40}\;/\\0.8675$\pm$0.0783}
		& \makecell{33.28$\pm$\best{1.13}\;/\\0.9255$\pm$0.0293}
		& \makecell{25.45$\pm$\best{1.12}\;/\\0.6693$\pm$0.0949}
		& \makecell{25.46$\pm$\best{1.22}\;/\\0.7616$\pm$0.0701}
		& \makecell{-\;/ \\ 287} \\
		\midrule
		
		\makecell{DRLF \\ (TPAMI'22) }
		& \makecell{43.21$\pm$\secondbest{2.29}\;/\\0.9811$\pm$\secondbest{0.0046}}
		& \makecell{39.05$\pm$\secondbest{1.45}\;/\\0.9800$\pm$\secondbest{0.0056}}
		& \makecell{40.11$\pm$2.61\;/\\0.9677$\pm$0.0061}
		& \makecell{35.76$\pm$1.74\;/\\0.9643$\pm$0.0099}
		& \makecell{36.13$\pm$\secondbest{2.75}\;/\\0.9303$\pm$0.0143}
		& \makecell{31.47$\pm$1.80\;/\\0.9237$\pm$0.0225}
		& \makecell{1673\;/\\2.50} \\
		\midrule
		
		\makecell{MSP \\ (TCI'23) }
		& \makecell{41.35$\pm$2.83\;/\\0.9771$\pm$0.0054}
		& \makecell{34.23$\pm$3.09\;/\\0.9682$\pm$0.0117}
		& \makecell{39.19$\pm$2.83\;/\\0.9641$\pm$0.0070}
		& \makecell{33.14$\pm$2.41\;/\\0.9533$\pm$0.0150}
		& \makecell{36.00$\pm$2.85\;/\\0.9335$\pm$0.0159}
		& \makecell{30.70$\pm$2.04\;/\\0.9183$\pm$0.0276}
		& \makecell{1200\;/\\3.74} \\
		\midrule
		
		\makecell{PFE \\ (IJCV'24) }
		& \makecell{42.81$\pm$2.33\;/\\0.9771$\pm$0.0053}
		& \makecell{37.90$\pm$1.70\;/\\0.9773$\pm$0.0066}
		& \makecell{40.23$\pm$2.60\;/\\0.9683$\pm$0.0064}
		& \makecell{35.39$\pm$1.71\;/\\0.9641$\pm$0.0111}
		& \makecell{36.49$\pm$2.78\;/\\0.9346$\pm$0.0140}
		& \makecell{31.73$\pm$\secondbest{1.75}\;/\\0.9281$\pm$0.0231}
		& \makecell{3236\;/\\12.94} \\
		\midrule
		
		\makecell{HLFRN \\(TCI'25)} 
		&
		\makecell{\secondbest{43.66}$\pm$2.36\;/\\\secondbest{0.9827}$\pm$\secondbest{0.0046}}
		&
		\makecell{\secondbest{40.11}$\pm$1.43\;/\\\secondbest{0.9830}$\pm$\secondbest{0.0056}}
		&
		\makecell{\secondbest{40.94}$\pm$2.65\;/\\\secondbest{0.9722}$\pm$\secondbest{0.0059}}
		&
		\makecell{\secondbest{37.23}$\pm$1.71\;/\\\secondbest{0.9717}$\pm$\secondbest{0.0086}}
		&
		\makecell{\secondbest{36.97}$\pm$2.87\;/\\\secondbest{0.9404}$\pm$\secondbest{0.0128}}
		&
		\makecell{\secondbest{33.14}$\pm$1.93\;/\\\secondbest{0.9396}$\pm$\secondbest{0.0204}}
		& \makecell{1676\;/\\15.26} \\
		\midrule
		
		\makecell{VSANet\\ (Ours)} 
		& \makecell{\best{44.08}$\pm$\secondbest{2.29}\;/\\\best{0.9835}$\pm$\best{0.0044}}
		& \makecell{\best{40.20}$\pm$\secondbest{1.39}\;/\\\best{0.9833}$\pm$\best{0.0055}}
		& \makecell{\best{41.31}$\pm$\secondbest{2.58}\;/\\\best{0.9732}$\pm$\best{0.0057}}
		& \makecell{\best{37.26}$\pm$\secondbest{1.66}\;/\\\best{0.9719}$\pm$\best{0.0084}}
		& \makecell{\best{37.34}$\pm$2.85\;/\\\best{0.9432}$\pm$\best{0.0127}}
		& \makecell{\best{33.21}$\pm$1.90\;/\\\best{0.9408}$\pm$\best{0.0195}}
		& \makecell{1142\;/\\10.79} \\
		
		\bottomrule
	\end{tabular}
	\label{tab:sota}
\end{table*}

\subsection{View-Aware Sparse Attention Block}
\label{subsec:vsa}

The key observation underlying the proposed VSA block is that noise is generally independent across LF views, whereas the underlying scene content exhibits strong spatial-angular correlation. Consequently, a noisy feature can be restored by aggregating similar features from other views, even when they are spatially distant. To exploit this property, we represent the entire 4D LF feature map as a unified token sequence and perform sparse non-local attention over all spatial-angular locations simultaneously.

Given an LF feature map
$\mathbf{L}_f\!\in\!\mathbb{R}^{S\times T\times H\times W\times C
}$, two parallel convolution layers are applied in the SS to generate query ($\mathbf{Q}$), key ($\mathbf{K}$), and value ($\mathbf{V}$) features. Specifically, a $3\!\times\!3$ convolution projects the features to dimension $C_r\!=\!C/r$ producing the query and key embeddings, while a $1\!\times\!1$ convolution generates value features of dimension $C$. The resulting features are reshaped using R1D operation into a unified token representation of length $N\!=\!S\cdot\!T\cdot\!H\cdot\!W$, yielding $\mathbf{Q},\mathbf{K}\!\in\!\mathbb{R}^{N\times C_r}$ and $\mathbf{V}\!\in\!\mathbb{R}^{N\times C}$. Unlike existing LF methods that restrict attention to spatial or angular subspaces \cite{many2many}, this formulation places all spatial-angular positions into a common attention space, enabling direct feature interaction across all views and and spatial locations.

Computing full pairwise attention over $N$ tokens incurs a complexity of $\mathcal{O}(N^2)$, which is prohibitive for LF data. To address this issue, we employ LSH-based sparse attention \cite{lsh,reformer}. The core idea of LSH is to assign similar tokens to the same hash bucket with high probability. Since a single hash round may occasionally fail to group correlated tokens due to the randomness of the projection, we perform $(n_h$ independent hash rounds (HRs) and aggregate their outputs \cite{reformer}. For $k^{th}$ HR, a random rotation
matrix $\mathbf{R}^{(k)}\!\in\!\mathbb{R}^{C_r\times\lfloor
B_h/2\rfloor}$ is generated, where $B_h$ denotes the number of hash buckets.
Each query/key token $\mathbf{q}_i/\mathbf{k}_i$ is projected onto the rotated space and assigned to a bucket according to,
\begin{equation}
  b_i^{(k)} = \arg\max\!\left(
    \bigl[\mathbf{R}^{(k)}\mathbf{q}_i;\,
    -\mathbf{R}^{(k)}\mathbf{q}_i\bigr]\right).
  \label{eq:lsh}
\end{equation}
Concatenating the projected feature with its negation improves bucket discrimination and reduces spurious token grouping. Because the same random rotation is applied to all $N$ tokens, similar features from different spatial-angular locations are likely to be assigned to the same bucket, thereby enabling implicit cross-view interaction.

Within each HR, tokens are first sorted according to their bucket indices and then partitioned into non-overlapping chunks of size $s$.
To alleviate boundary effects, each chunk is augmented with its two neighboring chunks, forming a $3s$-token context window. Scaled dot-product attention is then computed within each local window as,
\begin{equation}
  a_{ij}^{(k)} =
  \frac{\exp\!\left(\mathbf{q}_i\cdot\hat{\mathbf{k}}_j
    \right)}
       {\sum_{\ell}\exp\!\left(\mathbf{q}_i\cdot
         \hat{\mathbf{k}}_\ell\right)},
  \quad
  \hat{\mathbf{k}} = \tfrac{\mathbf{k}}{\|\mathbf{k}\|},
  \label{eq:attn}
\end{equation}
yielding the attended output
$\mathbf{z}_i^{(k)}\!=\!\sum_j a_{ij}^{(k)}\mathbf{v}_j$,
where $\mathbf{v}_j$ is the value token.
This reduces per-round complexity from $\mathcal{O}(N^2)$
to $\mathcal{O}(N\cdot s)$.
Finally, the outputs from all HRs are combined using confidence-weighted aggregation:
\begin{equation}
  \mathbf{z} = \sum_{k=1}^{n_h}\pi_k\,\mathbf{z}^{(k)},
  \qquad
  \pi_k =
  \frac{\exp\!\left(\log \mathbf{Z}^{(k)}\right)}
       {\sum_{k'}\exp\!\left(\log \mathbf{Z}^{(k')}\right)},
  \label{eq:multiround}
\end{equation}
where $\log \mathbf{Z}^{(k)}\!=\!\log\sum_j\exp(\mathbf{q}_i\cdot
\hat{\mathbf{k}}_j)$ represents the log-sum-exp normalization term of the attention scores and serves as a confidence measure for the corresponding HR. Consequently, rounds that discover stronger feature correspondences contribute more to the final representation. The aggregated feature $\mathbf{z}$ is reshaped back to
$\mathbb{R}^{S\times T\times H\times W\times C}$
and added to $\textbf{L}_f$ through a residual connection.

\section{Experiments}
\label{sec_4}

\subsection{Experimental Setup}
\label{setup}
We use two benchmark datasets for evaluation. Following \cite{hlfrn}, we use $70$ image pairs for training and $30$ image pairs for testing from the STFLytro dataset \cite{stflytro}. We also test on $10$ image pairs from the EPFL dataset \cite{epfl} without any fine-tuning to check the model's generalization ability. Following \cite{hlfrn}, the center $5\times 5$ SAIs are used for both datasets. Zero mean additive white Gaussian noise (AWGN) with standard variance of $\sigma= 10, 20, 50$ were
synthesized to generate the noisy images.
We train VSANet with Adam optimizer for $10,000$ epochs with a batch size of $4$. The initial  learning rate is $2\times10^{-4}$, which is halved in every 2500 epoch. In each iteration, we randomly crop the SAIs to a patch size of $32\times 32$. We use the $L_1$ loss function between the denoised and ground truth (GT) images for training the network.
 All experiments are conducted using an NVIDIA A40 GPU within the PyTorch framework. We compare denoising performance with mean and standard deviation (SD) of PSNR and SSIM values of all SAIs. A higher mean and lower SD indicate better performance.

In VSANet, the number of SEAR modules is set to $G\!=\!5$, each containing $F\!=\!5$ SFE blocks. In the channel attention layer of CBAMs, we employ pointwise convolution and the channel reduction factor is $16$. For the spatial attention layer of CBAM in the SS, the convolution kernel size is set to be $7\times 7$, and for other convolutions,  kernel size is set to be $3\times 3$. The feature dimension is set to $C\!=\!32$. In each VSA block, we set the reduction factor $r\!=\!2$, hash bucket size $B_h\!=\!128$, chunk size $s\!=\!64$, and perform $n_h\!=\!6$ HRs.

\begin{figure*} 
    \centering
  \includegraphics[width=0.92\linewidth]{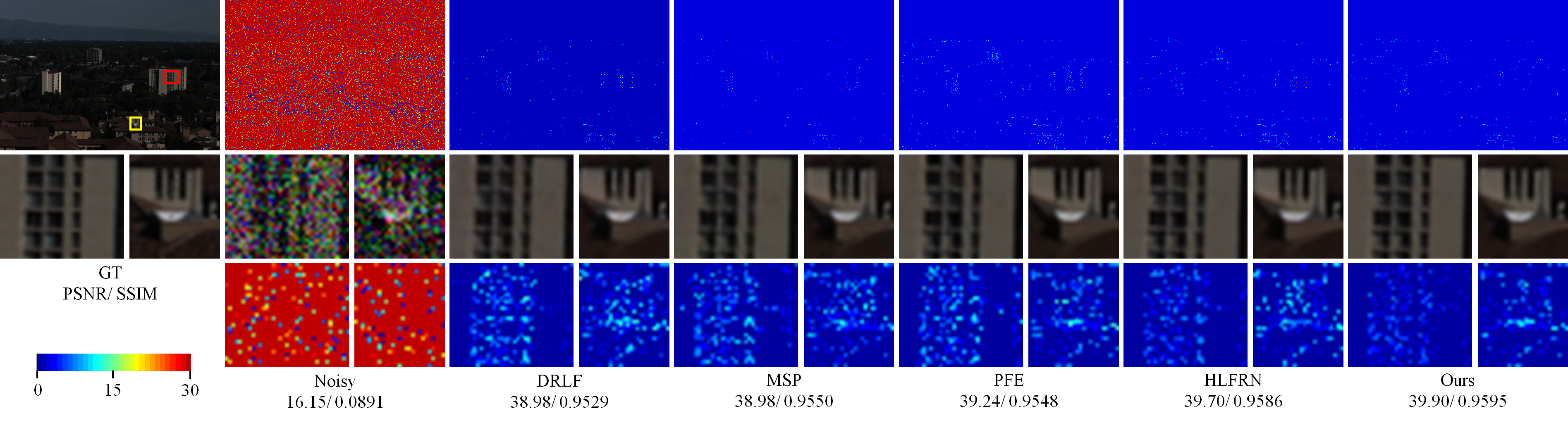}
  \includegraphics[width=0.92\linewidth]{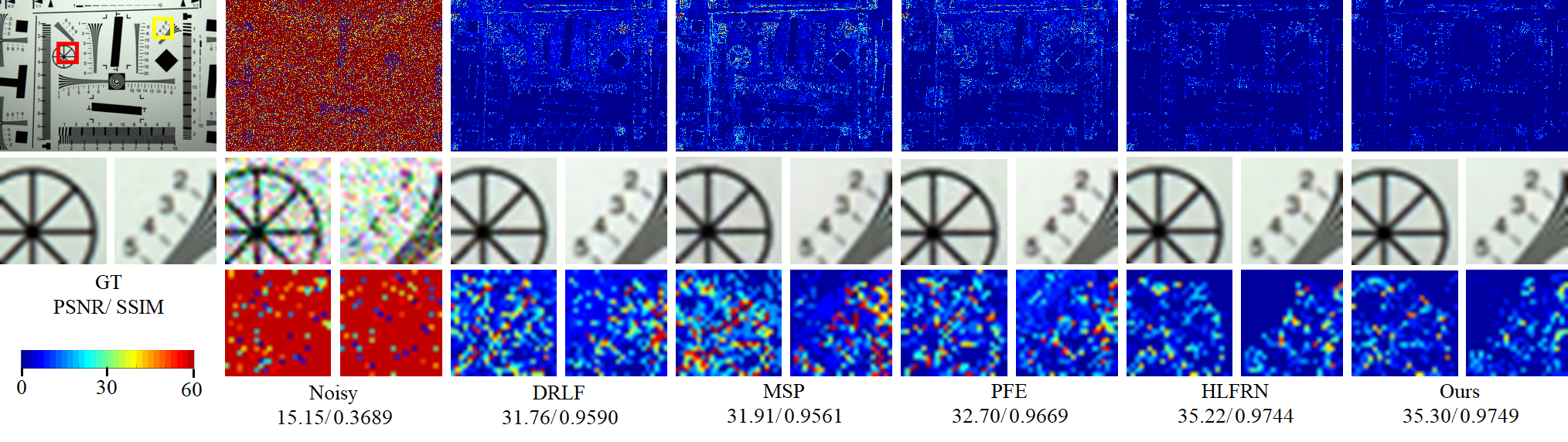}
  \caption{Visual comparison with SOTA methods. Top: STFLytro dataset, bottom: EPFL dataset. Best viewed at $400\%$ zoom.}
  \label{fig:sota} 
\end{figure*}
\subsection{Comparison with the SOTA Methods}
\label{sota}
VSANet is compared with one traditional method, LFBM5D \cite{lfbm5d}, and four recent DL-based methods: DRLF \cite{drlf}, MSP \cite{msp}, PFE \cite{pfe}, and HLFRN \cite{hlfrn}. Table \ref{tab:sota} reports quantitative results on two datasets across three noise levels. In addition to PSNR and SSIM, we include model parameters and average runtime, measured for denoising an LF image of resolution $532\times364$. VSANet achieves the best mean PSNR and SSIM across all noise levels on both datasets, and also attains the highest SSIM on SD. While HLFRN shows comparable performance, VSANet requires significantly fewer parameters and lower runtime.

Fig. \ref{fig:sota} shows visual comparison on one image from STFLytro dataset and one image from EPFL dataset for noise level $\sigma=50$. We show the center view of the LF images. Compared to the SOTA methods, VSANet better removes the noise and preserves the structural details.

\begin{table}[t!]
	\centering
	\caption{Comparison of results on ablation experiments (AEs). We highlight the best performance in \best{red} color, and underline the second-best performance in \secondbest{blue} color.}
	\begin{tabular}{r|cc}
		\toprule
		AE &  \makecell{STFLytro\\ PSNR \;/\;SSIM} & \makecell{EPFL\\ PSNR \;/\;SSIM} \\
		\midrule
		AE1 & \makecell{36.79$\pm$\best{2.78}\;/\;0.9382$\pm$0.0141}
		& \makecell{32.47$\pm$\secondbest{1.78}\;/\;0.9336$\pm$0.0207} \\
		AE2 & \makecell{37.11$\pm$\secondbest{2.79}\;/\;0.9418$\pm$0.0132}
		& \makecell{33.02$\pm$\best{1.77}\;/\;0.9396$\pm$\secondbest{0.0193}} \\
		AE3 & \makecell{37.27$\pm$2.84\;/\;0.9425$\pm$0.0131}
		& \makecell{33.05$\pm$1.89\;/\;0.9395$\pm$0.0204} \\
		AE4 & \makecell{37.13$\pm$2.84\;/\;0.9413$\pm$0.0133}
		& \makecell{32.98$\pm$1.87\;/\;0.9385$\pm$0.0197} \\
		AE5 & \makecell{37.28$\pm$2.85\;/\;0.9415$\pm$0.0129}
		& \makecell{32.79$\pm$2.37\;/\;0.9385$\pm$0.0208} \\
		AE6 & \makecell{36.82$\pm$2.92\;/\;0.9393$\pm$0.0139}
		& \makecell{32.51$\pm$1.87\;/\;0.9358$\pm$0.0210} \\
		AE7 & \makecell{37.26$\pm$2.87\;/\;0.9426$\pm$0.0133}
		& \makecell{33.13$\pm$1.88\;/\;0.9400$\pm$0.0197} \\
		AE8 & \makecell{\best{37.36}$\pm$2.84\;/\;\best{0.9435}$\pm$0.0128}
		& \makecell{\best{33.23}$\pm$1.89\;/\;\secondbest{0.9407}$\pm$0.0199} \\
		AE9 & \makecell{37.31$\pm$2.84\;/\;0.9418$\pm$\best{0.0122}}
		& \makecell{33.18$\pm$1.87\;/\;\best{0.9408}$\pm$\best{0.0189}} \\
		AE10 & \makecell{\secondbest{37.34}$\pm$2.85\;/\;\secondbest{0.9433}$\pm$0.0132}
		& \makecell{33.18$\pm$1.87\;/\;0.9406$\pm$0.0200} \\
		Ours & \makecell{\secondbest{37.34}$\pm$2.85\;/\;0.9432$\pm$\secondbest{0.0127}}
		& \makecell{\secondbest{33.21}$\pm$1.90\;/\;\best{0.9408}$\pm$0.0195} \\
		\bottomrule
	\end{tabular}
	\label{tab:ablation}
\end{table}
\subsection{Ablation Study}
\label{ablation}
In this section, we conduct ablation experiments (AEs) to validate the effectiveness of our proposed method. Table \ref{tab:ablation} presents the quantitative comparison of the AEs on two datasets for noise level $\sigma=50$. 

To validate the effectiveness of VSA block, we conduct following three experiments. \textbf{AE1: }VSANet without VSA block, \textbf{AE2: }VSA block is designed to perform sparse attention within each view instead of the entire 4D LF, \textbf{AE3: }Instead of multi-round aggregation, one round of LSH is performed in the VSA block. As presented in Table \ref{tab:ablation}, all these AEs degrade the performance.

To verify the effectiveness of FR block, we conduct following two experiments. \textbf{AE4: }FR block is removed from the SEAR module, \textbf{AE5: }FR block is designed to refine features only in the SS. As shown in Table \ref{tab:ablation}, both these AEs degrade the performance.

The FE block extract features in the SS, AS, EHS, and EVS. To verify its effectiveness, we conduct \textbf{AE6} that designs FE block to extract features only in the SS. As exhibited in Table \ref{tab:ablation}, AE6 significantly degrades the performance.

We conduct two AEs to check the effects of changing number of SEAR modules in VSANet. \textbf{AE7: }$G$ is set to $4$, \textbf{AE8: }$G$ is set to $6$. Although $G=6$ yields slightly better performance, it also increases parameter count and runtime (params count/ runtime for \textbf{AE8} are 1370K/12.55s; for our setting, 1142K/10.79s). Therefore, we set $G=5$ in VSANet.
We conduct two AEs to check the effects of changing number of SFE blocks in the SEAR module. \textbf{AE9: }$F$ is set to $4$, \textbf{AE10: }$F$ is set to $6$. As shown in Table \ref{tab:ablation}, $F=5$ gives better performance.

 All the results in Table \ref{tab:ablation} justify the design of our proposed VSANet.

\section{Conclusion}
\label{sec_5}
We presented VSANet, a view-aware sparse attention network for LF image denoising. The VSA block exploits LF signal-noise asymmetry by representing the 4D LF as a unified spatial-angular token space and performing implicit cross-view aggregation through LSH-based sparse attention. Feature refinement across spatial, angular, and epipolar subspaces further enhances representation. Ablation studies validate the proposed architectural components. VSANet outperforms SOTA methods while maintaining favorable computational efficiency. Future work could extend the VSA framework to other LF restoration tasks.
\bibliographystyle{IEEEtran}
\bibliography{main}

\end{document}